\documentclass[fleqn,10pt]{wlscirep}
\usepackage[utf8]{inputenc}
\usepackage[T1]{fontenc}
\title{Model-based actor-critic: GAN (model generator) + DRL (actor-critic) => AGI} 

\author{Aras Dargazany (arasdar@uri.edu)}

\begin{abstract}
Our effort is toward unifying GAN and DRL algorithms into a unifying AI model (AGI or general-purpose AI or artificial general intelligence) which has general-purpose applications to: 
\textbf{(A) offline learning} (of stored data) like GAN in (un/semi-/fully-)SL setting such as big data analytics (mining) \& visualization; 
\textbf{(B) online learning} (of real or simulated devices) like DRL in RL setting (with/out environment reward) such as (real or simulated) robotics \& control;
Our core proposal is adding an (generative/predictive) environment model to the actor-critic (model-free) architecture which results in a model-based actor-critic architecture with temporal-differencing (TD) error and an episodic memory. 
The proposed AI model is similar to (model-free) DDPG and therefore it can also be called model-based DDPG.
To evaluate it, we compare it with (model-free) DDPG by applying them both to a variety (wide range) of independent simulated robotic and control task environments in OpenAI Gym and Unity ML Agents.
Our initial limited experiments show that DRL and GAN in model-based actor-critic results in an incremental goal-driven intellignce required to solve each task with similar performance to (model-free) DDPG.
Our future focus is to investigate the proposed AI model potential to:
\textbf{(A) unify DRL field} inside AI by producing competitive performance compared to the best of model-based (PlaNet) and model-free (D4PG) approaches;
\textbf{(B) bridge the gap between AI and robotics} communities by solving the important problem of reward engineering with learning the reward function by demonstration;
\end{abstract}

\begin{document}
\maketitle

\section*{Introduction: GAN (generative model) vs DRL (actor-critic) -- offline learning vs online learning}
Currently all problems in artificial intelligence (AI) are formulated as an end-to-end (deep) artificial neural network optimization problem, also known as deep neural network learning or in short deep learning (DL)\cite{lecun2015deep}.
These problems are mainly classified into: offline learning and online learning;
Offline learning refers to learning from the stored data or dataset which are nowadays known as big data problem.
In offline learning, the stored big or small dataset might not have any labels (unsupervised), might be partially labeled (semi-supervised) or might be fully labeled (fully-supervised).
Online learning refers to learning from devices or machines such as robots (simulated or real) directly in real-time to achieve a specific goal or accomplish a specific task.
Online learning might be refered to as reinforcement learning (RL) or reward-based learning or robot learning or robust control learning as well.
RL is sometimes refered to as deep RL (DRL) which is due to the application of DL to RL problems.
Originally DL\cite{lecun2015deep} was proposed as a powerful solution to offline learning, more specifically supervised learning (SL) of big data such as Imagenet classification\cite{krizhevsky2012imagenet} and TIMIT speech recognition\cite{hinton2012deep}.
Currently, generative adversarial nets (GAN)\cite{goodfellow2014generative} \& deep reinforcement learning (DRL)\cite{mnih2015human}, specifically actor-critic (AC) algorithms are the fore-frontiers of DL/AI community:
(A) GAN for offline learning or (un/, semi-/,fully-)supervised learning (SL)\cite{kingma2014semi, salimans2016improved}; and 
(B) AC for online learning or (deep) RL or DRL\cite{lillicrap2015continuous, barth2018distributed}.
DL formulates an optimization problem with a user-defined objective (loss or cost) function and lacks a single unifying objective (loss or cost) function\cite{lecun2015deep,pfau2016connecting}.
GAN \& AC both consist of two models (networks): one of which approximates the main function for prediction (actor for AC or generator for GAN) and another one approximates the objective (loss or cost) function (discriminator for GAN or critic for AC).
This has revolutionized the assumptions behind DL algorithms in terms of having a pre-defined fixed objective (loss or cost) function although applying gradient descent (and ascent) to these networks (GAN \& AC) often leads to mode collapse, unstable training and no convergence in some cases. 
AC methods\cite{barto1983neuronlike, sutton2000policy} and GAN\cite{goodfellow2014generative} are two main
classes of multilevel (in fact bilevel) optimization problems with close connection\cite{pfau2016connecting}.
Since both of these hybrid or multilevel optimization models suffer from stability issues, mode collapse, and convergence difficulties, techniques\cite{pfau2016connecting}
for stabilizing training have been developed largely independently by the two communities.
This work is mainly inspired and built upon some important previous works directly or indirectly~\cite{lotfi2009combining, dolatabadi20142d, dolatabadi2014optimized,gyllinsky2017anear, dargazany2009hand, dargazany2009kernel, dargazany2010multibandwidth, dargazany2012human, dargazany2014stereo, dargazany2014terrain, dargazany2019iterative, dargazany2018wearabledl, dargazany2019deep, dargazany2019end, dargazany2019human,dargazany2019mlr, dargazany2019stereo}. 
In this work, we mainly focus on the two most stable one of each: Wasserstein GAN (WGAN)\cite{arjovsky2017wasserstein} and deep deterministic policy gradients (DDPG)\cite{lillicrap2015continuous}.
The main contribution of this work is to point out:
1. the strong connection/similarity between WGAN and DDPG as the two most stable classes of GAN and AC (DRL) model;
2. Given the first contribution, we argue that DDPG demonstrates adversarial learning behavior very similar to WGAN;
3. Given the first contribution, how this similarity can lead to a model-based DDPG (model-based actor-critic);
Eventually, we want to conclude that adversarial intelligence (as the product of adversarial learning process/behavior) might be a general-purpose AI (AGI) since it is applicable to both offline learning through GAN for (un-, semi-, fully-)SL and online learning through AC (DRL) for RL and robot learning. 

\section*{GAN for offline learning} 
Generative adversarial networks (GAN)\cite{goodfellow2014generative} formulates the unsupervised learning problem as a game between
two opponents: a generator G which generates sample images from a random noise sampled from a fixed probability distribution/density function (PDF) or a fixed noise source such as normal function;
and a discriminator D which
classifies the sample images as real/true or fake/false (maps an image to a binary classification probability); 
The original or vanilla GAN\cite{goodfellow2014generative} was formulated as a zero sum game with the cross-entropy loss
between the discriminator prediction and the true identity of the image as real or generated/fake.
To make sure that the generator has gradients from which to learn even when the discriminator classification accuracy is high, the generator loss function is usually formulated as maximizing the probability of classifying a generated sample as true rather than minimizing its probability of being classified false (or fake).
GAN is formulated in a bilevel optimization setting.

\subsection*{GAN for offline learning or (un, semi-, fully-) supervised learning (learning from data)}
GAN was initially applied to unsupervised learning (no label available for the dataset) for image generation, data distribution modelling and discovering the data structure\cite{goodfellow2014generative}.
Later on, it was applied to (semi-)supervised learning (SL) problem\cite{kingma2014semi, salimans2016improved} where the data is either partially labeled (semi-supervised) or fully labeled (fully-supervised)\cite{pix2pix2016} such as image classification, object detection, and object recognition.
The first application of GAN to (semi-)SL
problem\cite{kingma2014semi}
reported 64\% accuracy on SVHN\footnote{street-view house numbers (~\href{http://ufldl.stanford.edu/housenumbers/}{http://ufldl.stanford.edu/the-street-view-house-numbers/})\label{svhn}} dataset with 1\% labeled data.
In another original work\cite{salimans2016improved} by Tim Salimans at OpenAI, GAN reaches over 94\% accuracy on the same SVHN dataset using only 1,000 labeled examples out of almost 100,000 examples which means only 1\% of the entire dataset is labeled.
Though originally GAN was proposed as a form of generative model for unsupervised learning\cite{goodfellow2014generative}, GAN has proven useful for semi-supervised learning\cite{salimans2016improved}, and fully supervised learning\cite{pix2pix2016}. 
Therefore, GAN is a potentially capable approach for offline-learning of big data such as big data analytics/mining and visualization.

\subsection*{GAN has no proof of convergence.}
Unfortunately, there is no proof of convergence for GAN.
No one has yet shown that GAN\cite{goodfellow2014generative} will converge to an equilibrium.
On contrary, very many practical experiences have shown that GAN is unstable and might lead to mode collapse which means technically no convergence (equilibrium) at all.
One ray of hope is provided in the idea of using Earth Mover’s metric (or Wasserstein metric)\cite{arjovsky2017wasserstein}.
MIX-GAN\cite{arora2017generalization} architecture with multiple generators and discriminators can also reach an approximate equilibrium under fairly tight conditions which are not usually practical.
Generative multi-adversarial network\cite{durugkar2016generative} shows that multiple discriminators does indeed improve empirical performance.

\subsection*{Wasserstein GAN: stable GAN with Wasserstein loss}
WGAN\cite{arjovsky2017wasserstein, salimans2016improved} uses Wasserstein loss (also known as earth mover) as an alternative loss function instead of the traditional/original GAN loss function, cross-entropy loss\cite{goodfellow2014generative}. 
Wasserstein loss improves the stability of GAN training/learning considerably by almost eliminating the important problem of mode collapse.
WGAN\cite{salimans2016improved} is applied to semi-supervised learning for classification without labels (partially labelled).
Using Wasserstein distance loss and some other tricks (gradient penalty), WGAN\cite{salimans2016improved} achieved state-of-the-art results in semi-supervised classification on MNIST, CIFAR10 and SVHN at the time of publication. 
Later on, we show that Wasserstein loss provides deep connection and deep similarity with Bellman loss (TD loss using Bellman equation or Q-learning or Q-loss learning) which is used in DDPG\cite{lillicrap2015continuous} as the most stable AC method in DRL.

\subsection*{Self-attention GAN: GAN using attention layers}
Self-Attention GAN (SAGAN) uses attention\cite{vaswani2017attention} for long-range dependency modeling (proven by\cite{ramsauer2020hopfield}) for video prediction~\cite{wang2018non} and image generation tasks. 
In contrast with vanilla GAN which can generate high-resolution images with only spatially local points, SAGAN can generate higher resolution images using spatially non-local points (or global and local points at the same time as mentioned in non-local neural net\cite{wang2018non}). 
Visualization of the attention layers in the generator shows the non-local regions which correspond to the object shapes are detected compared to the local regions of a fixed shape in the vanilla GAN.

\section*{DRL (actor-critic) for online learning (real-time learning by interacting with environment)}
Actor-critic (AC) methods are a long-established class of techniques in reinforcement learning (RL) [2].
While most RL algorithms either focus on learning a value function like value iteration
and TD-learning (value-based methods), 
or learning a policy directly such as policy gradient methods (policy-based methods),
AC methods learn
both the actor (policy) and the critic (value) simultaneously. 
In some AC methods, the actor model (policy function) is
updated with respect to the approximate critic model (value function), in which case learning behavior and architecture similar to those in GAN can result.
%
Formally, consider the typical Markov decision processes (MDP) setting for RL, where we have a set of states S, actions A, and discount factor [0, 1].
The aim of AC methods is to simultaneously learn an action-value function 
that predicts the total expected discounted future reward
and learn a policy that is optimal for that value function.
Actor-critic is also formulated in a bilevel optimization setting similar to GAN:
There are many AC methods that attempt to solve this problem. 
The distinction between these methods lies mainly
in the way training proceeds. 
Traditional AC methods optimize the
policy through policy gradients and scale the policy gradient by Bellman loss, also known as temporal-difference (TD) loss or TD error, while the action-value
function is updated by ordinary TD error learning\cite{sutton1988learning}.
In this work, we mainly focus on deep deterministic policy gradients (DDPG)\cite{lillicrap2015continuous} which
is intended for the case where actions and
observations are continuous, and use deep learning (DL) for function approximation of both the action-value function (critic) and the state-action function or policy (actor).
DDPG is an established DRL (AC) approach with continuous actions which updates the policy (actor) by backpropagating the gradients of the estimated value from critic with respect to the actions rather than backpropagating the TD error directly.

\subsection*{Actor-critic for online learning and (deep) RL or DRL}
Deep Q-learning (DQL) or deep Q-network\cite{mnih2015human} was one of the breakthrough work in RL and the beginning of deep RL (DRL).
Actor-critic (AC) algorithms\cite{crites1995actor} are one of the most powerful RL or DRL algorithms which are composed of two networks: actor and critic.
AC methods are models from deep reinforcement learning (DRL) in which a model learns an action-value function that predicts the expected total future reward (the Critic), and a policy that is optimal for maximizing that value (the Actor or controller).
DDPG\cite{lillicrap2015continuous}, as a stable DL-based AC method, is in fact the extension of DQL approach to the continuous action space since DQL was in fact applied to the Atari game for discrete action space control and search.

\subsection*{Actor-critic methods have proof of convergence to an optimal policy.}
The original AC method 
\cite{barto1983neuronlike} can be viewed as the formal beginning of work in computational reinforcement learning. 
The critic in AC is like the discriminator in GAN, and the actor in AC methods is like the generator in GAN.
In both systems, there is a game being played between the actor (generator) and the critic (discriminator). 
The actor begins exploring the state space, and the critic should learn to evaluate the random exploratory behavior of the actor.
It was formally proved by Konda\cite{konda2000actor, konda2002actor, konda2003onactor} that in any Markov decision process (MDP), AC methods will eventually converge to the optimal policy.
The proof was non-trivial and shown a decade after AC proposal\cite{barto1983neuronlike}.
Recently (35 years after AC methods proposal) Google DeepMind proposed the same AC algorithm combined with DL (DDPG\cite{lillicrap2015continuous}) to solve the difficult Atari games from raw video input.

\subsection*{Deep Deterministic Policy Gradients: stable AC (DRL) method with Bellman/TD loss}
Deep Deterministic Policy Gradients (DDPG)\cite{lillicrap2015continuous} (stable AC approach in terms of convergence) is the application of DL to AC methods.
DDPG uses Bellman loss\cite{mnih2015human} or temporal-difference (TD) loss\cite{sutton2000policy} as the loss function.
TD loss learning is also know as Q-learning\cite{sutton2000policy} which after the introduction of DL\cite{lecun2015deep} gave birth to Deep Q-learning (DQL) or deep Q-network\cite{mnih2015human} which was the beginning of deep RL (DRL).
DDPG\cite{lillicrap2015continuous}, as a stable DL-based AC method, is in fact the extension of DQL approach to the continuous action space since DQL was in fact applied to the Atari games for discrete action space control and search.

\section*{Connection between GAN and AC methods}
Both GAN and AC methods can be seen as bilevel or two-time-scale optimization problems, meaning one model is optimized with respect to the optimum of another model.
Bilevel optimization problems have been extensively studied under AC methods by Konda\cite{konda2000actor, konda2002actor, konda2003onactor} mainly under the assumption that both optimization problems are linear or convex\cite{colson2007overview}.
On contrary, both DDPG\cite{lillicrap2015continuous} and WGAN\cite{arjovsky2017wasserstein} which are the center of attention in this work and the most stable kind of GAN and AC methods, are in fact non-linear and have non-convex optimization surfaces due to DL which are non-linear function approximator.
The question is though:
(A) if GAN and AC are by nature the same algorithms or not?
Or (B) can one emerge from another? if yes, which emerges from which?
Pfau and Vinyals\cite{pfau2016connecting} show that GAN can be viewed as AC methods in an environment where the actor cannot affect the reward.
Therefore, they answer the above questions as following:
(A) Yes, they are connected;
(B) Yes, GAN can emerge from AC methods;
They review a number of extensions to GAN and AC algorithms with even more complicated information flow. 
They encourage both GAN and AC (DRL) communities to develop general, scalable, and stable algorithms for multilevel optimization with DL, and to draw inspiration across communities.
Pfau and Vinyals\cite{pfau2016connecting} confirmed that AC and GAN are siblings and therefore there must be a parent algorithm/architecture that include both methods.
Pfau and Vinyals\cite{pfau2016connecting} encourage us for more investigation of deeper connection (and possibly convergence and unification) between GAN and AC methods for the development and adoption of a more general-purpose AI technique (AGI) applicable as GAN or AC (DRL).
%
Pfau and Vinyal\cite{pfau2016connecting} describes the connection between GAN and AC as an MDP in which GAN is a modified version of AC method as following:
Consider an MDP where the actions set every pixel in an image.
The environment randomly chooses either to show the image generated by the actor or show the real image.
Let the reward from the environment be 1 if the environment chose the real image and 0 if not. 
This MDP is stateless as the image generated by the actor does not affect future data.
The AC architecture learning in this environment resembles the GAN game.
A few adjustments have to be made to make it identical.
If the actor had access to the state, it could trivially pass a real image forward.
Therefore the actor must be a blind actor, with no knowledge of the state.
Stateless MDP doesn't prevent the actor from learning though.
The mean-squared Bellman/TD loss is usually used as the loss function for AC (specifically DDPG), mean-cross entropy adversarial loss is used instead for GAN (so-called GAN loss).
The actor's parameters in AC should not be updated if the environment shows a real image. 
Critic zeros down its gradients for the actor (no update for actor's parameters/weights) if the reward is 1 for the real image.
If the reward is 0 for the fake image, critic still zeros down its gradients for the actor. 
If the reward is 1 for the fake image, it's time for updating actor's parameters using critic's output and gradients.
This is how, GAN can be seen as a modified AC method with blind actor in stateless MDP.

\subsection*{Is actor-critic adversarial?}
According to \cite{pfau2016connecting}, it is not obvious why an AC algorithm should lead to an adversarial behavior;
Typically the actor and critic are trying to optimize complimentary loss functions (compatible or orthogonal), rather than optimize the same loss function in different directions (adversarial). 
The adversarial behavior in AC emerges due to the stateless MDP in which the GAN game is being played since the actor cannot have any causal effect on the reward in this stateless MDP.
A critic, however, cannot learn the causal structure of the game from
input examples alone, and moves in the direction of features that predict reward more accurately (minimizing the reward prediction error/loss). 
The actor moves in a direction to increase reward (maximizing the reward value) based on the best estimate from the critic, 
but this change cannot
lead to an increase in the true reward, so the critic will quickly learn to assign lower value in the direction the actor has moved.
Thus the updates to the actor and critic, which ideally would be
orthogonal (as in compatible actor-critic or complimentary loss functions) instead becomes adversarial. 
Despite these differences between GAN and typical RL problems (settings), we believe there are enough similarities to generalize between GAN and AC (DRL) algorithms.
This generalization might be the path toward general-purpose AI or artificial general intelligence (AGI).

\subsection*{Target networks is the only main difference between AC method (specifically DDPG) and GAN (specifically WGAN).}
Since the action-value function appears twice in the Bellman equation, 
stability can be an issue in Q-learning with function approximation.
Target networks
address this by fixing one of the networks in TD updates or possibly slowing down the updates of this network so-called target network.
This will turn the Q-learning problem from RL to SL and helps the Q-learning to converge whereas without it will most likely diverge.
Since the GAN game can be seen as a stateless MDP, the second appearance of the action value function disappears.
Therefore, 
we do not consider target networks applicable to the GAN setting\cite{pfau2016connecting}.

\subsection*{Compatibility of actor-critic}
According to \cite{pfau2016connecting},
one of the unique theoretical developments of AC methods is the
notion that the actor and critic are compatible or complementary in terms of loss function.
It is not
clear if the notion of compatibility can be naturally extended to the GAN setting. 
We would generally prefer our GAN to be adversarial than compatible.
We hope that by pointing out the deep connections between GAN and AC (DRL), we encourage both communities of GAN and DRL (AC) to merge and join forces.

\section*{Model-based actor-critic: GAN + DRL (actor-critic)}
We propose to teach machines to accomplish
more complex tasks in one common environment with modelling the real world/environment with rich textures and complex structural compositions.
We address three challenges in particular for training an agent to model
the real-world:
\textbf{First}, how to apply the adversarial learning~\cite{goodfellow2014generative} to improve the quality of the generated environment model ($\tilde{env}$) since GAN is proved to be effective in image generation and modelling data distribution tasks~\cite{ledig2017photo}.
This is also known as explicit modelling of the environment. 
Actor-critic (AC) methods, specifically deep deterministic policy gradients (DDPG)~\cite{lillicrap2015continuous} models the environment internally in the critic network.
We chose DDPG as our default AC approach due to the continuous action space of the agent.
\textbf{Second}, how to build an efficient differentiable neural renderer
that can simulate the
environment and is transferable to other tasks as well so that we don't need to start training the environment from scratch.
Also the more tasks our agent accomplishes, the more accurate our environment model (the generator or the neural differentiable simulator/renderer) becomes.
We train a generative neural network which directly maps the current action ($A$) and the current state of environment/world $S$ into the next state of environment ($S^{'}$).
This differential renderer (generator) can be
combined with AC and turn into model-based AC that
can be trained in an end-to-end fashion, which might significantly
boosts both the modelling quality and convergence speed in terms of solving the task.
\textit{NOTE: generator, neural simulator, differential renderer, environment modeller, and world renderer are all the same.}
\textbf{Third}, how to design a reward function is another really important challenge since reinforcement learning (RL or deep RL also refered as DRL) or reward-based learning is impossible without the reward.
Manually designing this reward function for the complex problems/tasks is almost impossible.
Specifically, if we want to apply DRL or deep RL to the robotics problem, we have to be able to design an accurate reward function or go for a simple task which the reward function design is fairly easy.
The latter is not always possible and therefore we have to be able to also learn the reward function.
This has turned into a headache for many researchers in artificial intelligence (AI) and robotics.
In summary, our contributions/solutions for the three aforementioned challenges/problems are also three-fold:
\textbf{1.} We approach the modelling task with the GAN and build a model-based DRL agent by combining GAN with (model-free) AC method (DDPG) that can model the agent's environment. 
To this end, we build a differentiable neural renderer (generator) for efficiently
modelling the environment. 
This neural environment renderer (generator) and a discriminator model the world by training model-based DRL agent in an end-to-end fashion.
Discriminator is required to discriminate between the real environment samples and fake ones (generated/rendered ones) by neural environment model.
\textbf{2.} Explicitly modelling the environment with generator neural net compared to implicitly modelling it in critic of AC method (DDPG~\cite{lillicrap2015continuous}) helps transferring this model to other problems and tasks in that same environment.
This transfer learning makes our DRL agent very sample efficient, meaning it can accomplish the same task with much less episodes compared to (model-free) AC or original DDPG.
\textbf{3.} Given the explicit transferable environment model (generator) and the target image (an image of the goal or what we want to accomplish/gain at the end), we can learn how to design the reward function with the discriminator as it becomes more and more powerful in discriminating between the real environment and explicit environment model.
This is also solves the problem of designing/hand-engineering the reward function for complex tasks.

\section*{Literature}
SPIRAL~\cite{ganin2018synthesizing} is an adversarially trained DRL
agent that learn the structures in images and tries to paint them from scratch, but fails to recover the
details of images.
SPIRAL++~\cite{mellor2019unsupervised} is an improved version of SPIRAL which is using GAN and RL agent at the same time in a reinforced adversarial learning fashion for paining an image in a canvas from scratch.
Both SPIRAL and SPIRAL++ are composed of an actor-critic and discriminator but they use a fixed/non-differentiable renderer which is generating the painting.
StrokeNet~\cite{zheng2018strokenet} 
combines differentiable neural image/paining renderer (a generator) on top of actor-critic and discriminator using recurrent neural network (RNN) to train agents to paint but fails to generalize on color images. 
The most inspiring and similar work to us is done by Huang et al.~\cite{huang2019learning} which is proposing a model-based DDPG which is in fact a model-based actor-critic as well.
He is proposing four neural networks in his model-based DDPG agent to learn how to paint on canvas from scratch and without any external reward.
He eventually experimentally compared his model-based DDPG to the original (model-free) DDPG and experimentally demonstrates the superiority of model-based DDPG.

\subsection*{DRL approaches}
Superficially, SPIRAL and SPIRAL++ can be seen as a model-free RL techniques for stroke
based, non-photorealistic rendering/generating.
Similarly to some modern stroke-based rendering techniques,
the positions of strokes are determined by a learned system (namely, an RL agent). 
Unlike traditional methods however, the objective function that determines the goodness of the output image is learned
unsupervised via the adversarial objective (adversarial loss). 
This adversarial objective allows us to train without access to ground-truth strokes,
enabling the method to be applied to domains where labels are prohibitively expensive to collect, and
for it to discover surprising and unexpected styles.
In this line of work, there are a number of works that use constructions similar to SPIRAL and SPIRAL++ to tune the parameters of non-differentiable/fixed simulators. 
In another line of work, Frans and Cheng (2018)~\cite{frans2018unsupervised}, Nakano (2019)~\cite{nakano2019neural}, Zheng et al. (2019)~\cite{zheng2018strokenet}, and Huang et al. (2019)~\cite{huang2019learning} achieve remarkable learning speed by relying on backpropagation/ neural model of the renderer, however they rely on being able to train an accurate model of the environment in the first instance.

\subsection*{Model-free vs model-based RL approaches}
Recently, several works (Frans and Cheng in 2018~\cite{frans2018unsupervised}, Nakano in 2019~\cite{nakano2019neural}, Zheng et al. in 2019~\cite{zheng2018strokenet}, and Huang et al. in 2019~\cite{huang2019learning}) proposed to improve sample efficiency and convergence stability of generative RL agents by replacing the actual non-differentiable/fixed rendering simulator with a differentiable neural one (generator neural net) which is trained offline to predict how an action affects the state of the canvas environment. 
Although promising, there are three scenarios in which model-free approaches like
SPIRAL and SPIRAL++ might be a better approach:
\textbf{Firstly}, one of the main advantages of the neural environments is training agents
by directly backpropagating the gradient coming from the objective of interest (either reconstruction
loss if there is no discriminator or adversarial loss if there is a discriminator or GAN setting). 
Unfortunately, this means that the action space of the agent has to be continuous to efficiently use backpropagation gradient (neural net learning). 
This means if the action space is discrete, we might have problem in learning our neural networks in model-based approaches.
\textbf{Secondly}, the success of the RL agent training largely depends on the quality of the neural model of the
environment. 
The simulator used in SPIRAL and SPIRAL++  is arguably easy to learn since new strokes interact with the existing drawing in a fairly straightforward manner. 
Environments which are highly stochastic or require handling of object occlusions and lighting might pose a challenge for neural network based environment models.
\textbf{Thirdly}, while in principle possible, designing a model for a simulator with complex dynamics may
be a non-trivial task. 
The majority of the recent works relying on neural renderers assume that the
simulator state is fully represented by the appearance of the canvas and therefore only consider
non-recurrent state transition models. 
There is no need for such an assumption in the SPIRAL and SPIRAL++ framework. 

\subsection*{Conclusion of the literature}
The use of neural simulators in model-based DRL are largely orthogonal to SPIRAL and SPIRAL++
and are likely to improve their performance further.
Nonetheless model-based DRL approaches are dependent on a given target image which is not the case for SPIRAL and SPIRAL++.

\section*{Model-based actor-critic}
\begin{figure}[ht!]
    \centering
    \includegraphics[width=\textwidth]{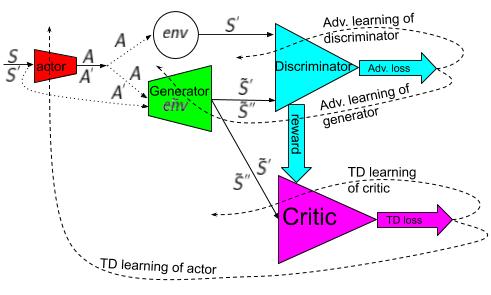}
    \caption{The overall architecture of model-based actor-critic (AC): 
    (a) At the inference/testing stage, the actor outputs an action ($A$) to the real environment ($env$) and generator ($\tilde{env}$).
    In result, the real environment ($env$) outputs the next state ($S^{'}$) to the actor and to the discriminator.
    The discriminator applies adversarial learning on Discriminator and Generator based on the resulting adversarial loss and update the environment model ($\tilde{env}$) more accurately.
    (b) At the training stage, the actor and critic are being trained/updated together using critic's temporal-difference (TD) loss through reinforcement learning (Q-learning). 
    The required reward for reinforcement learning (Q-learning) is given by the discriminator at each step, and the training samples are randomly sampled from the replay buffer.
    Adversarial learning (GAN learning) is also being done the same as inference/testing stage.
    $S$: current states of the environment; $S^{'}$: next states of the environment; $A$: actions; $A^{'}$: next actions, $env$: real environment/world of the agent; $\tilde{env}$: environment model which is being modelled by generator; $\tilde{S}^{'}$: predicted/generated states after current states of environment; $\tilde{S}^{''}$: predicted/generated states after next states of environment; Adv. learning: adversarial learning; TD loss: temporal-difference loss;}
    \label{fig:mbac}
\end{figure}

We propose the model-based DRL framework shown in Figure~\ref{fig:mbac}.
This framework is mainly composed of two stream frameworks: GAN framework for adversarial learning stream and AC framework for reinforcement learning stream.

\subsection*{GAN framework for adversarial learning stream}
\begin{figure}[ht!]
    \centering
    \includegraphics[width=\textwidth]{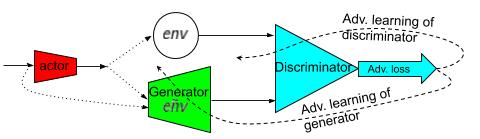}
    \caption{Adversarial learning stream in model-based actor-critic architecture}
    \label{fig:mbacAL}
\end{figure}

Since GAN has been widely used because of its great ability in modelling the data distribution by measuring the distribution distance between the generated (fake) data and the target (real) data, therefore the adversarial learning stream is mainly composed of GAN, a generator and discriminator, with an actor as you can see in figure~\ref{fig:mbacAL}.
Wasserstein GAN (WGAN)~\cite{arjovsky2017wasserstein} is our preference compared the original GAN~\cite{goodfellow2014generative} since it is an improved version of
the original GAN~\cite{goodfellow2014generative} which uses the $Wasserstein-l$ distance, also
known as Earth-Mover distance. 
The objective of the discriminator in WGAN is defined as~\ref{eq:wganobjective}:
\begin{equation}
    \max_{D}\mathbb{E}_{y\sim \mu }[D(y)]-\mathbb{E}_{x\sim \nu }[D(x)]
    \label{eq:wganobjective}
\end{equation}
where $D$ denotes the discriminator, $\nu$ and $\mu$ are the fake
samples and real samples distribution.
The adversarial learning stream is responsible for two important tasks:
\textbf{1.} modeling the environment using the generator as accurately as possible.
Discriminator in this stream helps discriminating between the real environment and the environment model $\tilde{env}$.
Using a generator neural network to model the environment has two advantages:
\textbf{First}, it is transferable to any other tasks in that same environment and also the more tasks we do with this environment model the more accurate it will become in terms of predicting the next state of the environment $st+1$ given the current state $st$ and current taken action $at$; 
\textbf{Second}, it boosts the performance of the agent in terms of successful completion of the task and sample-efficiency which in simple words means completing the task in much less episodes of try-and-error;
The training samples can also be generated randomly using Computer Graphics rendering programs.
The generator (environment model $\tilde{env}$) can be quickly trained with supervised learning on the GPU/s.
The model-based transition dynamics $st+1 = \tilde{env}(st, at)$ and the reward function $r(st, at)$ are differentiable neural network.
The generator is a deep (end-to-end) neural network consisting of
several fully connected layers and convolution layers. 

\begin{figure}[ht!]
    \centering
    \includegraphics[width=\textwidth]{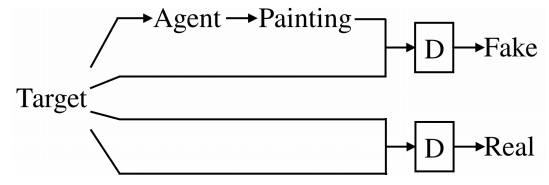}
    \caption{This is the training paradigm for discriminator according to Huang~\cite{huang2019learning}.}
    \label{fig:rewardlearning}
\end{figure}

\textbf{2.} learning the reward function for the reinforcement learning stream since the reinforcement learning stream is impossible without the the reward function.
Using a discriminator neural network to model the reward has also two advantages:
\textbf{First}, it is adaptable to other tasks in that same environment and we don't have to manually engineer/design the reward function;
\textbf{Second}, it boosts the performance of the agent in terms of successful completion of the task by accurately learning the appropriate reward function for that specific task;
The discriminator similar to generator is also a deep (end-to-end) neural network consisting of
several fully connected layers and convolution layers but the output is a scalar value/score $L(st)$. 
The conditional GAN training schema~\cite{isola2017image} is suggested by~\cite{huang2019learning} as the reward metric, where fake samples and the target vs real samples and the target are used for reward learning as shown in Figure~\ref{fig:rewardlearning} taken from~\cite{huang2019learning}. 
we want to reduce the differences between the current state and
target state as much as possible.
To achieve this, we set the difference of discriminator scores from $st$ to $st+1$ using equation~\ref{eq:rewardloss}
as the reward for guiding the learning of the actor.
\begin{equation}
    r(st, at) = L(st) - L(st+1)
    \label{eq:rewardloss}
\end{equation}
\subsection{Actor-critic framework for DRL}
\begin{figure}[ht!]
    \centering
    \includegraphics[width=\textwidth]{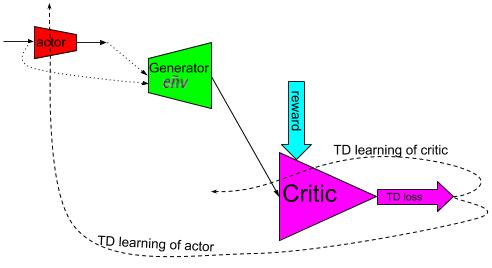}
    \caption{Reinforcement learning stream in model-based actor-critic architecture}
    \label{fig:mbacRL}
\end{figure}

The reinforcement learning stream is mainly composed of the actor-critic model and the generator which contain the model of the environment $\tilde{env}$ as shown in figure~\ref{fig:mbacRL}.
The optimization of the RL agent using the model-based AC is different from that using the original AC.
At timestep $t$, the critic takes $st+1$ as input instead of both of $st$ and $at$ as shown in figure~\ref{fig:diff}. 
The critic still predicts the expected reward $V(st)$ for the state $st$ but without the need for current action as the input also shown in figure~\ref{fig:diff}.
The new expected reward $V(st+1)$ is a value function $V(st)$ trained using discounted reward:
$V(st) = r + \gamma V(st+1)$
Here $r(st, at)$ is the reward when performing action $at$ based on $st$.
The environment function $st+1 = \tilde{env}(st, at)$ is the generator network which models the environment.

\section*{Model parameters and variables}
We model a task as an MDP with a state space $S$ an action space $A$, a reward function $r(st, at)$, an environment function $env(st, at)$ which can be the real environment $env$ or the environment model $\tilde{env}$. 
The details of these components are as following:
\textbf{The state space} includes all information regarding the state of the environment which can be observed by the agent.
\textbf{The environment function} $st+1 = env(st, at)$ makes the transition process from the current state of the environment to the next state.
\textbf{The action space} $at$ of the agent is a set of continuous parameters that control the position and orientation of the agent or part of the agent required for completing the target task.
We define the behavior of an agent as a policy function $\pi$ that maps states to deterministic actions ($\pi: S \rightarrow A$). 
\textbf{Timesteps} $t$ is every step that the agent make an observation of the environment state $st$ and takes an action accordingly $at$ and in result of the taken action, the current state of the environment evolves/transitioned based on the environment function $st+1 = env(st, at)$ which can be the real one $env$ or the fake one $\tilde{env}$.
\textbf{Reward} Selecting a suitable metric to measure the difference between the current state and the target state is crucial for RL agent.
In fact, reinforcement learning is reward-based learning and without a stable reward function/metric it is not possible. 
The reward is defined as follows: $r(st, at) = L(st) - \gamma L(st+1)$
where $r(st, at)$ is the reward at step $t$, $L(st)$ is the measured loss between the target environment state $I$ and the current state $st$ and $L(st+1)$ is the measured loss between the target environment state $I$ and the next state $st+1$.
$L(st)$ and $L(st+1)$ is formulated as the discriminator output score for current environment state $st$ and the next state$st+1$.
To reach the target environment state $I$, the agent should model the environment accurately for an precise prediction of the next state (adversarial learning stream using GAN) and maximize the cumulative rewards (reinforcement learning stream using AC) in one episode. 

\section*{Why model-based actor-critic?}
\begin{figure}[ht!]
    \centering
    \includegraphics[width=\textwidth]{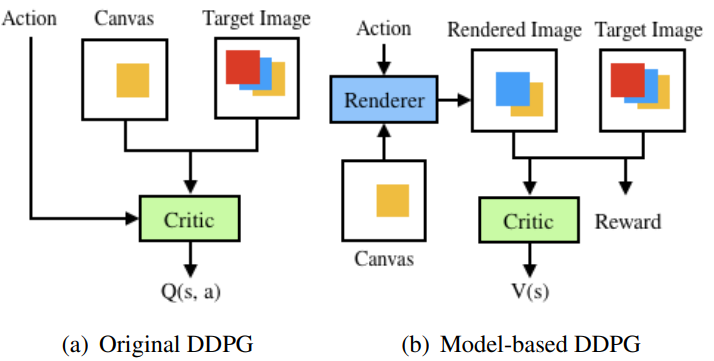}
    \caption{According to Huang~\cite{huang2019learning}: In the original actor-critic (DDPG~\cite{lillicrap2015continuous}), critic learns an implicit model of the environment.
    In the model-based actor-critic, the environment is explicitly modeled by a neural renderer (generator in our work $\tilde{env}$), which helps to train an agent efficiently.}
    \label{fig:diff}
\end{figure}

Since the action space in the control and robotic tasks are mainly continuous and high dimensional,
discretizing the action space to adapt to discrete DRL methods such as deep Q-network (DQN)~\cite{mnih2015human} and policy gradients (PG)~\cite{sutton2000policy} is quite burdensome if possible.
In contrast, deterministic policy gradients (DPG)~\cite{silver2014deterministic} uses deterministic policy to resolve the difficulties caused by high-dimensional continuous action space.
If the environment states are also high-dimensional, (original) DDPG~\cite{lillicrap2015continuous} solves this problem too using deep (end-to-end) convolutional neural network (DNN) so-called deep learning (DL)~\cite{lecun2015deep, krizhevsky2012imagenet}.
In the original DDPG, there are two networks: the actor $\pi(s)$ and critic $Q(s, a)$.
The actor models a policy $\pi$ that maps a state $st$ to action $at$. 
The critic estimates the expected reward for the agent taking action $at$ at state
$st$, which is trained using Bellman equation~\ref{eq:bellmanEq} as in Q-learning~\cite{watkins1992q} and the data is sampled from an experience replay buffer:
\begin{equation}
    Q(st, at) = r(st, at) + \gamma Q(st+1, \pi(st+1))
    \label{eq:bellmanEq}
\end{equation}
Here $r(st, at)$ is a reward given by the environment when performing an action $at$ at the state $st$. 
The actor $\pi(st)$ is trained to maximize the critic’s estimated value $Q(st, \pi(st))$. 
In other words, the actor decides an action/policy for each state. 
Based on the current state and the target state, the critic predicts an expected reward for the action. 
The critic is optimized to estimate more accurate expected rewards.
We cannot train a good-performance RL agent using original DDPG because it’s hard for the agent to model the complex environment well composed of real-world images/observations during learning. 
The World Model~\cite{ha2018recurrent} is a method to make agent understand the environments effectively. 
Similar to~\cite{huang2019learning, nakano2019neural, frans2018unsupervised}, we design a neural model of the environment (refered to as neural renderer in the cited works) so that the agent can model the environment effectively.
Then it can explore the modelled environment and improve its policy efficiently.
The difference between the two algorithms visually shown in Figure~\ref{fig:diff} from~\cite{huang2019learning} is very similar to what we are proposing.

\section*{Experiments: Initial limited experimental results}
\begin{figure}[ht!]
    \centering
    \includegraphics[width=\textwidth]{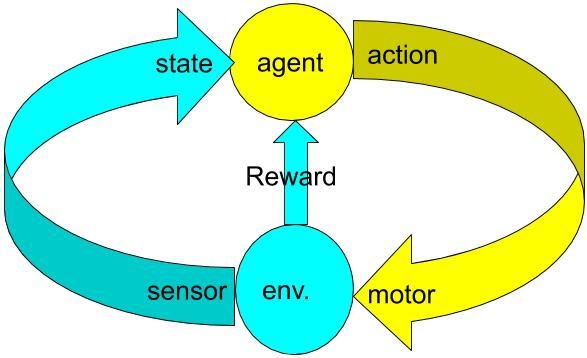}
    \caption{The inputs and outputs of the AI model and the task environment;
    The reward is how we can bridge the gap between agent \& environment;
    The reward is also how we can define the task/goal for the robot;
    agent: the proposed AI model which is based on model-based actor-critic, env: environment which is the task is being presented to the agent;}
    \label{fig:AI-experiments}
\end{figure}

We implemented and tested the proposed model-based actor-critic in some simulated environments such as OpenAI Gym and Unity ML agents that simulates a number of independent tasks in their own environment which provides both sensors input and the reward function as visualized in figure~\ref{fig:AI-experiments}. 
These two simulators provide a number of tasks in their own unique environments that varies from classical control problems (e.g. CartPole), robotic problems (e.g. Reacher arm), and famous video games (e.g. Car race).
These task environments are independent from each other in a sense that the knowledge from one is not required nor can be transferred to another one.
The inputs and output of the AI model (agent) and the task environment (env) are visually shown in figure~\ref{fig:AI-experiments} for better understanding of the data flow in between the AI model (agent) and the environment task (env) and how the experiments are being performed.

\subsection*{Reacher environment simulated by Unity ML-agents as a robotic arm problem}
\begin{figure}[ht!]
    \centering
    \includegraphics[width=\textwidth]{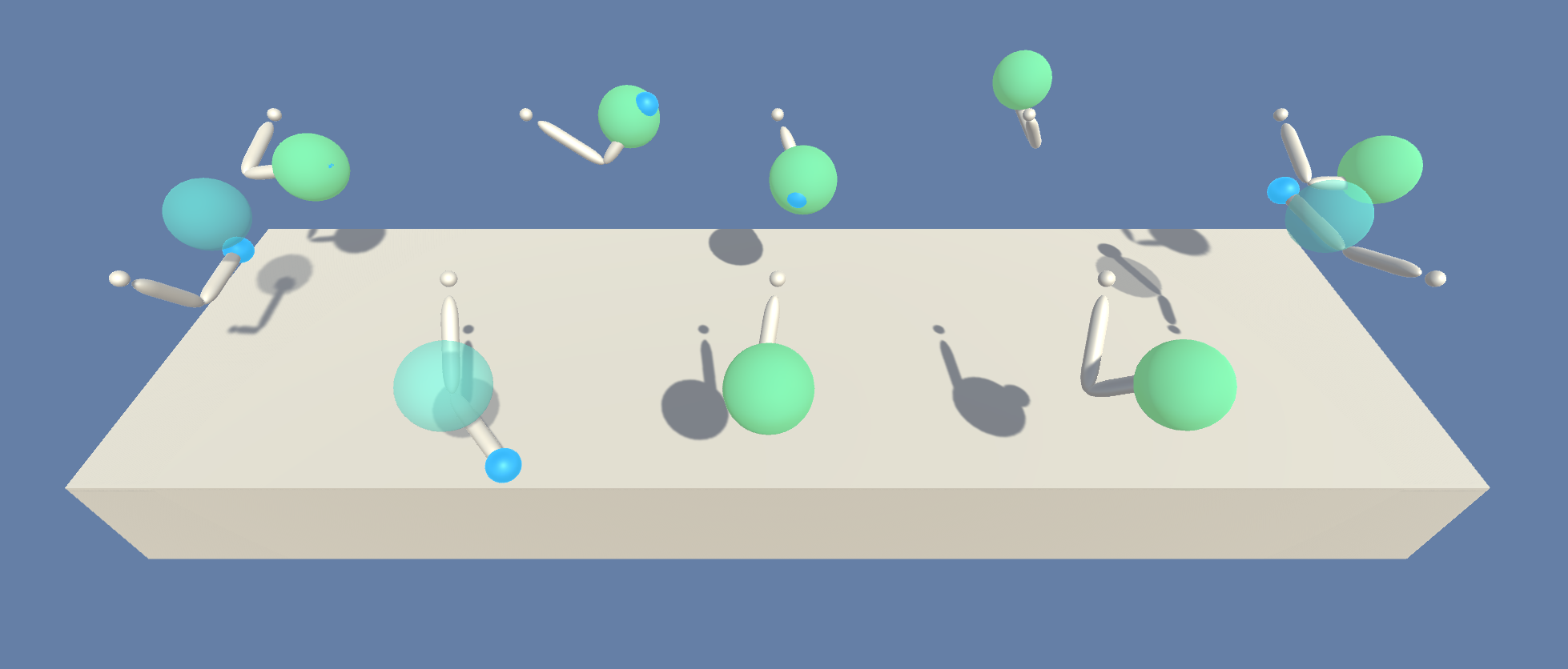}
    \caption{The Reacher environment in Unity, capable of using one or multi-agents experimental platform.}
    \label{fig:reacher}
\end{figure}

\begin{figure}[ht!]
    \centering
    \includegraphics[width=\textwidth]{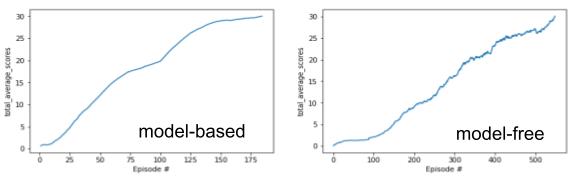}
    \caption{The current performance of the proposed approach in one Reacher and multiple Reachers environment;
    Y-axis is the total average score (average accumulated rewards) over the number of episodes on X-axis;
    We believe model-based actor-critic can reduce the number of episodes (improve sample efficiency) compared to original (model-free) actor-critic.}
    \label{fig:reacher-performance}
\end{figure}

Reacher is one of the Unity ML-agents environments for deep reinforcement learning (DRL) research experiments.
Reacher environment features and specifications are listed as following:
Reacher is a double-jointed arm which can move to target locations;
\textbf{Goal}: The agents must move its hand to the goal location, and keep it there;
\textbf{Agent Reward Function (independent)}: For each step, agent's hand reaches the goal location, it receives +0.1;
\textbf{State space:} A vector of 26 variables corresponding to position, rotation, velocity, and angular velocities of the two arm rigid bodies;
\textbf{Action space:} A vector of 4 continuous variables corresponding to torque applicable to two joints;
\textbf{To solve this task, Benchmark Mean Reward: 30}
Reacher environment is capable of using one agents or 20 agents (multi-agents) (figure~\ref{fig:reacher}).
%
We have implemented the proposed model-based actor-critic for initially experimentation on solving such tasks (figure~\ref{fig:AI-experiments}) using PyTorch library which is a python-based deep learning framework for Facebook.
The initial results of applying the model-based actor-critic to the Reacher environment with one agent or multi-agents (twenty) are shown in figure~\ref{fig:reacher-performance}.
Based on the solving criteria of Reacher, the proposed model-based actor-critic solved this task with one agent in roughly 500 episodes and with twenty agents (multi-agent) in roughly 175 episodes as shown in figure~\ref{fig:reacher-performance}.
We conducted this experiment to make sure that we can implement the proposed architecture or it is implementable.
We also wanted to make sure that it works in terms of solving reinforcement learning tasks.

\section*{Conclusion \& future perspective}
Our limited experiments show that deep reinforcement learning (DRL) and GAN in 
(our AGI model) can result in an incremental goal-driven intellignce required to potentially solve (general-purpose) variety of independent tasks, each in their own separate independent environments.
Our future focus is to investigate:
\begin{itemize}
    \item the connection between the model-based actor-critic (DDPG) and the brain: is model-based DDPG architecture and learning compatible and plausible with the brain?
    \item the connection between GAN and the brain: is there any adversarial learning in the brain?
    \item the application of GAN to (semi-/fully-) SL problems for offline learning of the stored data for big data analytics (mining) \& visualization: is it applicable to all variety of SL problems?
    \item the application of model-based actor-critic (DDPG) to variety of independent tasks in only one same environment with reward (or reward function) such as DeepMind control suite: can it transfer skill from one task to another?
    \item the application of model-based actor-critic (DDPG) for (simulated or real) robotic control without reward signal from the environment: can we learn reward function instead of manually engineering/designing it in a robotic environment?
\end{itemize}

\subsection*{Connection between model-based actor-critic and the brain}
\begin{figure}[ht!]
    \centering
    \includegraphics[width=\textwidth]{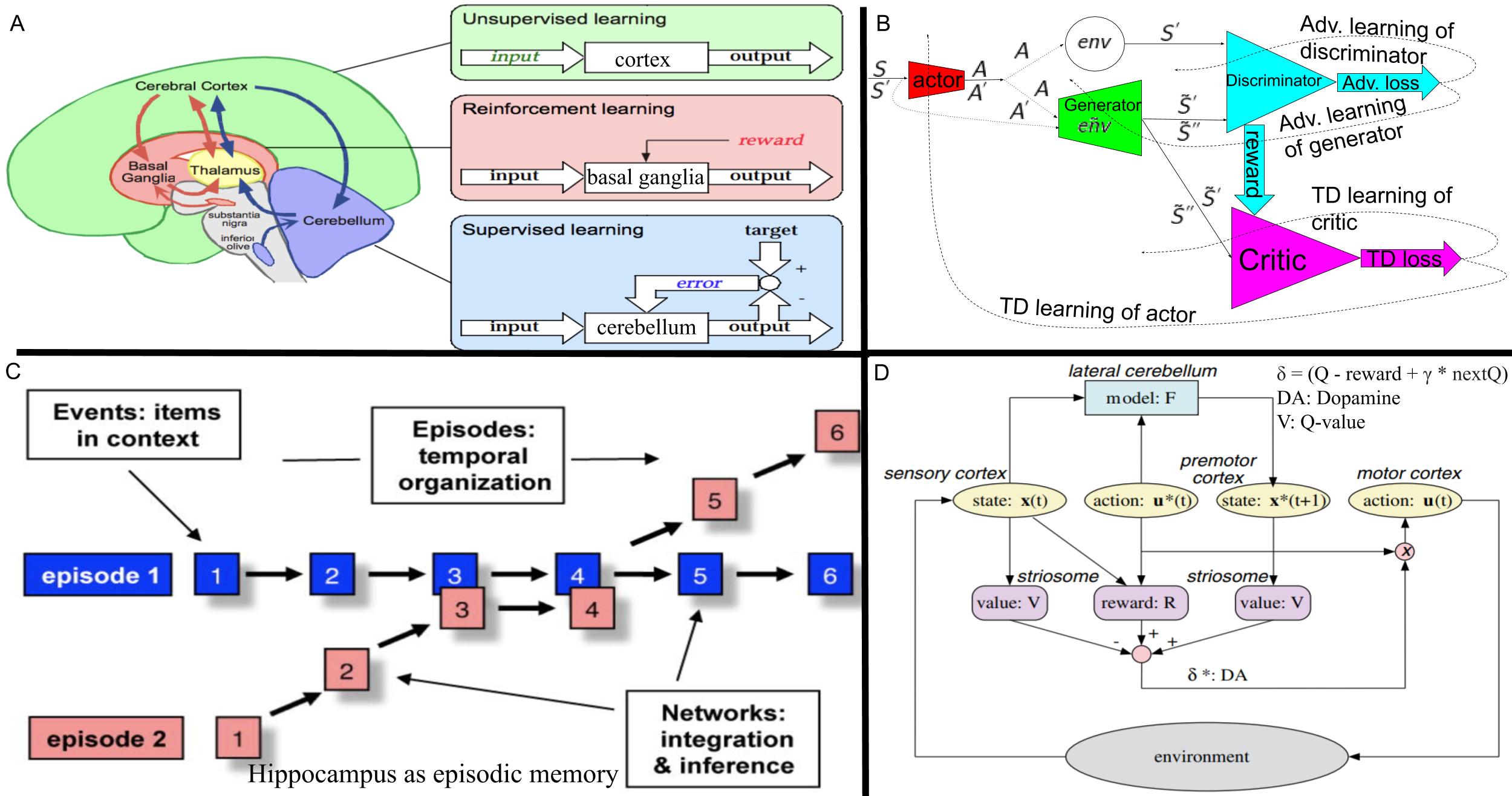}
    \caption{The connection between the model-based DDPG (actor-critic) and brain: 
    (A) the recent findings of the brain architecture (integrated network of BG, CBL, CTX, and hippocampus~\cite{caligiore2017consensus, bostan2018basal}) and brain learning paradigm (super-learning~\cite{caligiore2019super}); 
    (B) The proposed model-based actor-critic (DDPG) which might help us toward reaching a unifying AI model (AGI) which is more plausible and compatible with the brain; 
    (C) Hippocampus role and functionality in connection with this architecture biologically and artificially according to~\cite{eichenbaum2014can}; 
    (D) An old computational paradigm similar to our proposal according to Doya~\cite{caligiore2017consensus, doya1999computations} which is the computational architecture of model-based action selection combines forward models in the CBL and reward predictors in the BG proposed by Doya in 1999~\cite{doya1999computations};
    }
    \label{fig:brain}
\end{figure}

\begin{figure}[ht!]
    \centering
    \includegraphics[width=\textwidth]{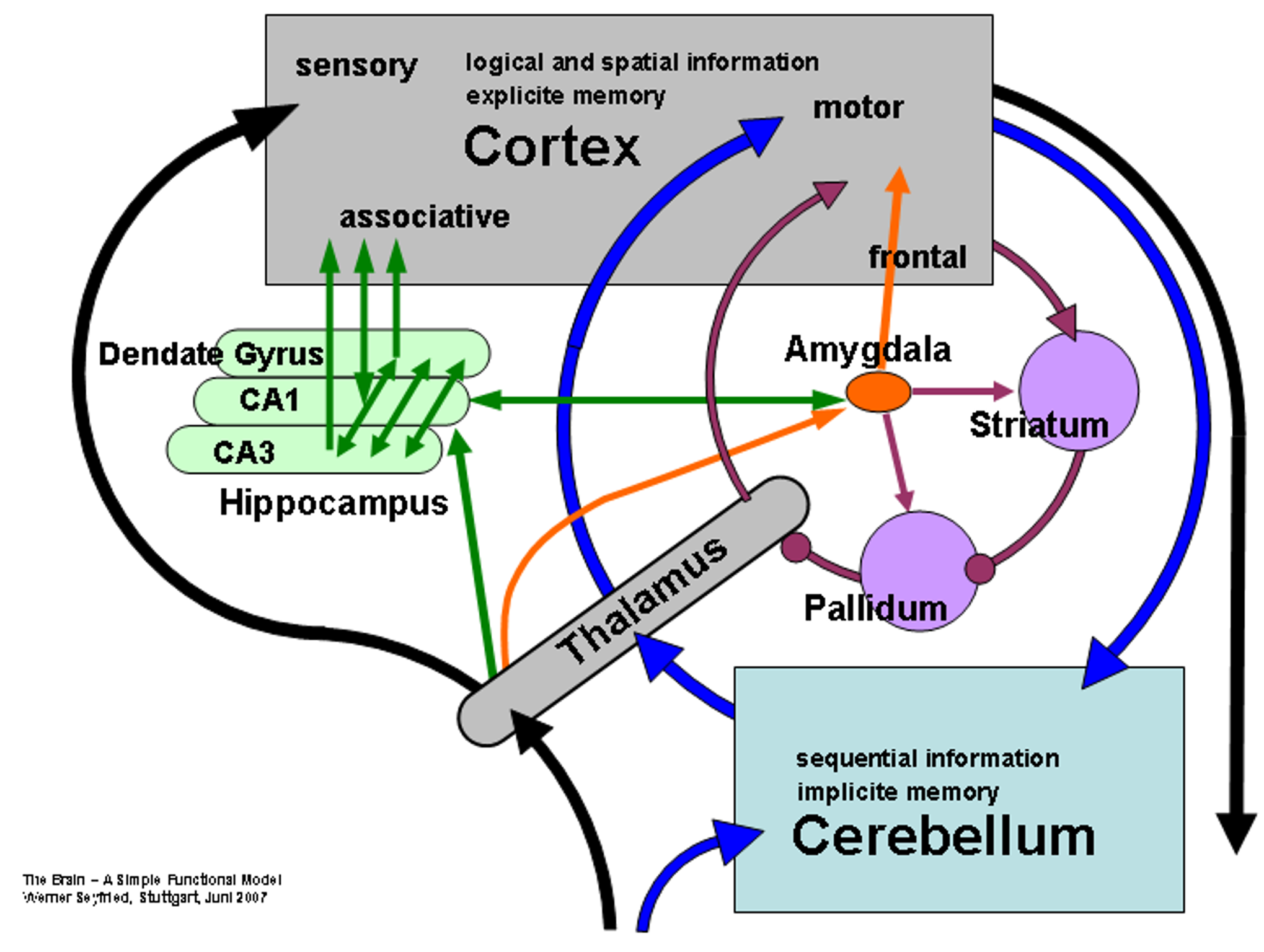}
    \caption{A similar old brain architecture proposed in this model~\cite{WWW-Brain-werner}}
    \label{fig:brain2}
\end{figure}

The proposed Model-based Actor-Critic algorithm/architecture is the most sensible/reasonable way of combining of GAN and DRL (AC algorithms).
I am reasoning that GAN can be used to model the environment/surroundings/our world as a generative model and in a realistic form.
And DRL (AC algorithms) can be used to to perform/come up with the set of actions which maximize the future reward in order to perform a task composed of multiple skills or accomplish a goal.
Neuroscientifically speaking, I highly believe this is how our brain functions or this is a systematic architecture of our brain (shown in figure~\ref{fig:brain} A): Cerebellum (CBL) can be the actor, Cortex (CTX) can be the generator to generate/model the environment and Basal ganglia (BG) can be the discriminator (indirect path) \& critic (direct path).
The work on this topic is very important in terms of building AGI models, combining model-free and model-based RL algorithms by figuring out how to combine GAN with DRL in a sensible/reasonable way.
A similar computational model of the brain is also illustrated in figure~\ref{fig:brain2}.

\subsection*{Hippocampus role in DRL systems}
Figure~\ref{fig:brain} C is a conceptual illustration of hippocampus as a short-term episodic memory space containing:
(A) Temporal events within a specific context in which they occurred. These episodes
are composed of temporal organization of events such as links between events and episodes;
(B) Spatial navigation or events as a spatial memory according to Eichenbaum (2004)~\cite{eichenbaum2014can};
According to this recent article~\href{https://singularityhub.com/2016/06/19/how-to-build-a-mind-this-learning-theory-may-hold-the-answer/}{https://singularityhub.com/how-to-build-a-mind-this-learning-theory-may-hold-the-answer/},
Kumaran et al.~\cite{kumaran2016learning} updates the complementary learning systems (CLS) theory proposed by and McClelland et al.~\cite{mcclelland1995there}, which holds that intelligent agents must possess two learning systems, instantiated in mammalians in neocortex and hippocampus.
The CLS theory states that the brain relies on two memory systems that allow it to rapidly soak in new information, while maintaining a structured model of the world that’s resilient to noise.
The core principles of the CLS theory is the understanding of memory in biological systems according to Kumaran et al..
In 1995, McClelland et al. observed a ground-breaking memory phenomenon in patients with damage to their hippocampus which was the fact that they could no longer form new memories although their past memories and concept were untouched and retrievable.
This led a land-mark paper by McClelland et al. in which they proposed CLS theory which in fact introduces hippocampus as a short-term memory (replay buffer or episodic memory) and cortex as long-term memory.
According to Bendor~\cite{bendor2012biasing}, the hippocampus does not fully replay all the recent activation patterns. 
Instead, it picks the most rewarding events and selectively replays them to the cortex.
This 'replay' has been postulated to be important for memory consolidation. 
Task-related cues can enhance memory consolidation when presented during a post-training sleep session, and, if memories are consolidated by hippocampal replay, a specific enhancement for this replay should be observed. 
Bendor~\cite{bendor2012biasing} also indicates that this replay during sleep can be manipulated by external stimulation which provide further evidence for the role of hippocampal replay in memory consolidation.
This might also means that rare but meaningful events might be prioritized for cortical learning, this is also known prioritized replay in DRL literature as well applied to D4PG~\cite{barth2018distributed} compared to DDPG~\cite{lillicrap2015continuous}.

\subsection*{Potential to unify DRL field inside AGI community}
Our proposed AGI model has the potential to unify DRL field inside AI community by producing competitive performance compared to the best of model-based (PlaNet\cite{hafner2018learning} and model-free (D4PG\cite{barth2018distributed}) approaches in DRL since our proposed AGI model architecture adds an environment model network on top of model-free architecture (DDPG as an AC model).
This environment model network functions as a long-term memory of environment since it models the environment and can predict the next state of the environment given its current state and the action applied to it.
Our proposed AGI model is based on policy network (actor), value network (critic), and model network (env. model or generator or generated env.) which makes it model-based, policy-based, and value-based all at the same time.
We should be able to compare our proposed AGI model with the best of model-based (PlaNet\cite{hafner2018learning} and model-free (D4PG\cite{barth2018distributed}) in the new DeepMind simulated environment (DeepMind Control Suite\cite{tassa2018deepmind}) which is built to benchmark model-based (and model-free) DRL algorithms.
This environment\cite{tassa2018deepmind} includes variety of tasks in one environment and under one environmental condition which makes it much more suitable to evaluate and compare DRL algorithms compared to OpenAI Gym and Unity ML-agents environments which are composed of variety of independent task environments with their own different independent environment conditions.

\subsection*{Towards learning the reward function by demonstration to bridge the gap between AI \& robotics}
The goal is learning-by-demonstration for the reward function problem or reward engineering problem.
The main problem in applying AI to robotics is the reward (and punishment) signal.
Hand-engineering the reward function for a goal-oriented/driven behavior is a very difficult and sometimes impossible problem.
Learning-by-demonstration\cite{singh2019end} helps us approach this problem using proposed AGI architecture.
The GAN network help us learn a task by the demonstration of a task\cite{singh2019end}.
Using Q-learning (and the Q-network or critic network or value network) 
As shown in figure~\ref{fig:mbac}, we can generate the reward by the discriminator network in every time-step for the robot similar to figure~\ref{fig:rewardlearning} proposed by Huang~\cite{huang2019learning}.
We have two different kinds of task for the robots: 
(a) the task we know how to do it so that we can demonstrate for the robot; 
(b) The task we don't know how to do it but we know how it looks like when it is done or accomplished;
Either one of them needs a reward function so that it can quantify the progress and the direction for the robot in terms of solving the task.
Therefore designing such reward function is the most crucial gap between robotics and AI.
Recently there have been some successful developments in the state of this art (reviewed by Amarjyoti\cite{amarjyoti2017deep}).
Our proposed AGI model tackles this problem or at least approaches this problem by learning the Q-network (critic network) which is related to defining reward function for the robot.
We can learn this reward function by demonstration of a user if the user knows how to demonstrate the task.
In case we don't know how to do the task, we should be able to provide some preliminary data to the AGI model so that it knows how the final state of environment looks like in terms success or failure.
The latter one sounds much more complicated than what we think it is and it needs much more investigation.


\end{document}